\documentclass{article}

\PassOptionsToPackage{numbers, compress}{natbib}
\usepackage[preprint]{neurips_2024}

\usepackage[utf8]{inputenc} % allow utf-8 input
\usepackage[T1]{fontenc}    % use 8-bit T1 fonts
\usepackage{hyperref}       % hyperlinks
\usepackage{url}            % simple URL typesetting
\usepackage{booktabs}       % professional-quality tables
\usepackage{amsfonts}       % blackboard math symbols
\usepackage{nicefrac}       % compact symbols for 1/2, etc.
\usepackage{microtype}      % microtypography
\usepackage{xcolor}         % colors
\usepackage{caption}
\usepackage{graphicx}
\usepackage{subfigure} 
\usepackage{multirow} 
\usepackage{adjustbox} 
\usepackage{arydshln}
\usepackage{bm}
\usepackage{algorithm}
\usepackage{algpseudocode}
\usepackage{amsmath}
\usepackage{wrapfig}
\usepackage{booktabs}
\usepackage{amsmath}
\usepackage{amssymb}

\title{Unlearning Information Bottleneck: Machine Unlearning of Systematic Patterns and Biases}

\author{
  Ling Han\\
  Yale University\\
  \And
  Hao Huang\\
  Wuhan University\\
  \And
  Dustin Scheinost\\
  Yale University\\
  \And
  Mary-Anne Hartley\\
  Yale University, EPFL\\
  \And
  María Rodríguez Martínez\\
  Yale University, University of Bern\\
  \And
  \texttt{\{l.han, dustin.scheinost, mary-anne.hartley,}\\\texttt{maria.rodriguezmartinez\}@yale.edu, haohuang@whu.edu.cn}
}

\begin{document}

\maketitle

\begin{abstract}
 Effective adaptation to distribution shifts in training data is pivotal for sustaining robustness in neural networks, especially when removing specific biases or outdated information—a process known as machine unlearning. Traditional approaches typically assume that data variations are random, which makes it difficult to adjust accurately the model parameters to remove patterns associated with the unlearned data. 
 In this work, we present the Unlearning Information Bottleneck (UIB), a novel information-theoretic framework designed to enhance the process of machine unlearning that effectively leverages the influence of systematic patterns and biases for parameter adjustment. 
 We propose a variational upper bound to recalibrate the model parameters through a dynamic prior that integrates changes in data distribution at an affordable computational cost, allowing efficient and accurate removal of outdated or unwanted data patterns and biases. 
 Our experiments across various datasets, models, and unlearning methodologies demonstrate that our approach effectively removes systematic patterns and biases while maintaining the performance of models post-unlearning. 
 % Our adaptive regularization strategy holds broad implications for more complex tasks, such as domain adaptation and continuous learning, thereby enhancing its practicality and scope.
\end{abstract}

\section{Introduction}
\begin{wrapfigure}{r}{0.5\textwidth}
  \begin{minipage}{0.5\textwidth}
    \centering
    \includegraphics[width=\textwidth]{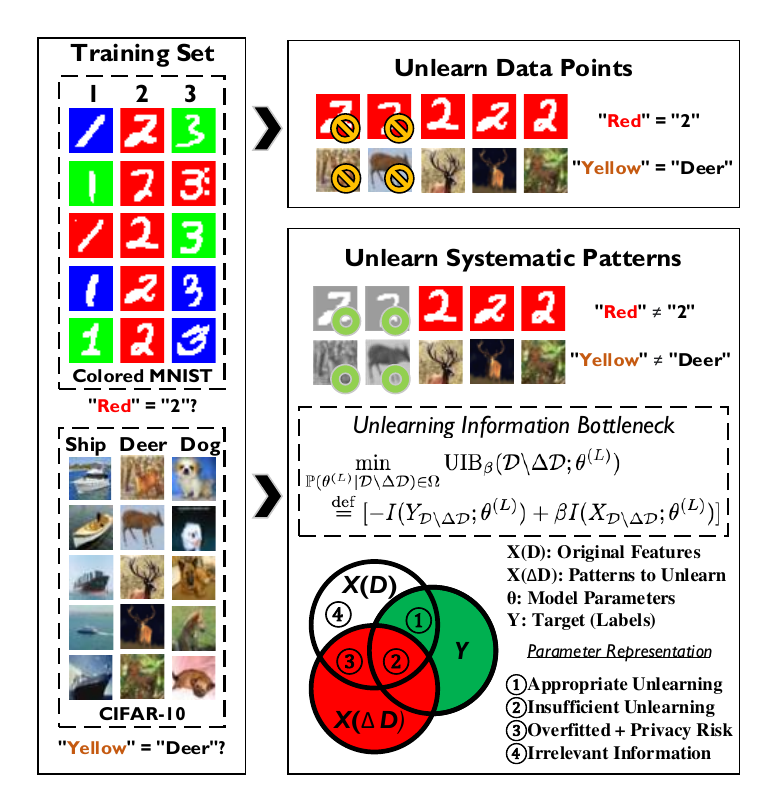}
    \captionsetup{font=small}
    \caption{The Unlearning Information Bottleneck (UIB) aims to optimize the parameter-space representation $\theta$. This process involves removing sufficient information $\Delta D$ that represents systematic patterns and biases from the original training dataset $D$, while simultaneously preserving the model's prediction capabilities for the target $Y$. The UIB framework ensures that $\theta$ avoids containing irrelevant information that could lead to overfitting, privacy breaches, or sensitivity to changes in model hyperparameters. Mutual information is denoted by I(·; ·), guiding the optimization to balance informativeness against redundancy.}
    \label{fig1}
  \end{minipage}
\end{wrapfigure}

While powerful, neural networks often retain traces of obsolete or sensitive information within their parameters, which can compromise data privacy and model integrity. The process of selectively removing such data influences from models, commonly called machine unlearning, is crucial for maintaining the trustworthiness and applicability of these systems in dynamic environments.
Current unlearning techniques predominantly operate under the assumption that changes to the training data occur randomly, which limits the precision with which models can be updated or corrected. This conventional approach fails to adequately address the structured nature of data variations that typically occur in real-world scenarios (As shown in Fig.~\ref{fig1}). 

The broader challenges of machine unlearning, particularly concerning systematic patterns and biases, have been addressed in multiple contexts, such as accuracy \cite{bourtoule2021machine}, timeliness \cite{cao2015towards}, and privacy \cite{chen2021machine}. However, the subtle complexities associated with unlearning features and labels in scenarios of non-random data removal require further examination. We delve into this nuanced landscape, illuminating the unique challenges inherent in unlearning systematic patterns and biases and proposing strategies to address them effectively.

In addressing the outlined challenges, we reconsider what constitutes an effective representation of unlearning. The Information Bottleneck (IB) framework provides a guiding principle for representation learning. The optimal representation should capture only the minimal sufficient information necessary for a downstream task. This principle encourages the representation to retain maximal predictive information relevant to the target (accuracy) while minimizing any additional, unnecessary information that could lead to overfitting (purging thoroughly). This selective approach ensures the unlearned model remains predictive and oblivious to unlearned data and biases.
Extending this IB principle to machine unlearning introduces new challenges. First, traditional methods leveraging IB rely on the assumption that the training data are independent and identically distributed (i.i.d.). However, this assumption does not hold for unlearning due to potential dependencies or correlations between data points. For instance, when specific data points are removed, shared features or biases violate the i.i.d. assumption. Additionally, data removal often introduces systematic patterns or biases, complicating the identification of minimal sufficient information. Developing an unlearning approach that extracts this relevant information while maintaining model performance is a challenge that necessitates thoughtful design.

\textbf{Proposed Work.} We present Unlearning Information Bottleneck (UIB), a novel information-theoretic framework designed to enhance the process of machine unlearning. Derived from the core principles of information bottleneck (IB), UIB adapts these concepts to manage systematic patterns and biases in data, ensuring that the learned representations remain minimal yet sufficient. By incorporating adaptive regularization, UIB recalibrates model parameters through a dynamic prior that integrates changes in data distribution, allowing efficient and accurate removal of outdated or unwanted data patterns.
To address the challenges posed by non-random data removal, UIB formalizes conditional distributions \(\mathbb{P}(\theta | \mathcal{D}\backslash\Delta \mathcal{D})\) in parameter space \(\Omega\) that follow a hierarchical structure to capture local dependencies among data points. This structure enables UIB to maintain effective parameter updates while minimizing the mutual information between model parameters and the dataset features, aligning the model's representation with privacy standards.

We derive variational bounds to make UIB practical and tractable, offering a variational upper bound that constrains the information from features while maximizing the predictive information in the representation. UIB is a robust solution for continual learning and unlearning, helping neural networks stay responsive and relevant in dynamically evolving data environments.
To this end, we develop an algorithm based on the UIB framework: UIB-IF, which utilizes an influence function to remove the impact of systematic features from the model. We generalized UIB-IF to other approximation methods by ensuring
the regularization term $\mathcal{R}^{(l)}$. 
Our experiments demonstrate that UIB enhances robustness compared to conventional unlearning approaches and consistently outperforms the sparsity and standard settings in mitigating biases effectively. 

The \textbf{contributions} of this paper are threefold:
(1) We introduce the Unlearning Information Bottleneck (UIB) framework, a novel approach designed to enhance the efficacy and performance of machine unlearning of patterns and biases;
(2) We derive variational bounds within the UIB framework that provide theoretical insights into the limits and capabilities of our unlearning methods;
(3) Through extensive experiments, we empirically validate the effectiveness of the UIB.

\section{Background}
\newcommand{\mD}{\mathcal{D}}
\newcommand{\mX}{\mathcal{X}}
\newcommand{\mY}{\mathcal{Y}}
\newcommand{\mI}{\mathcal{I}}
\newcommand{\mF}{\mathcal{F}}
\newcommand{\vf}{\mathbf{f}} 

We will initially revisit pertinent concepts from information theory and associated techniques in the parameter space for machine unlearning.

\paragraph{Problem Definition} Consider supervised learning problems where the original model is trained on a dataset $\mathcal{D}$ containing $N$ i.i.d. data realizations $\mathcal{D}=\left\{x^{(n)}, y^{(n)}\right\}_{n=1}^N=\left(x_{\mathcal{D}}, y_{\mathcal{D}}\right)$ of features $x \in \mathcal{X}$ and labels $y \in \mathcal{Y}$, within feature space $\mathcal{X} \subseteq \mathbb{R}^m$ and label space $\mathcal{Y} \subseteq \mathbb{R}^l$. We receive a machine unlearning request $\Delta \mathcal{D}$, which involves removing $N^{\prime}$ i.i.d. data realizations $\Delta \mathcal{D}=$ $\left\{x[i]^{(n)}, y[j]^{(n)}\right\}_{n=1}^{N^{\prime}}=\left(x_{\Delta \mathcal{D}}, y_{\Delta \mathcal{D}}\right)$ from the training dataset $\mathcal{D}$, where $i$ denotes the $i$-th feature of $x$ and $j$ denotes the $j$-th label of $y$. Define a pattern point \( z_{KL}^{(n)} \) where \( K \) is a set of feature indices and \( L \) is a set of label indices. Here, \( K = \{k_1, k_2, \ldots, k_p\} \) and \( L = \{l_1, l_2, \ldots, l_q\} \), with \( p \) and \( q \) representing the number of features and labels, respectively. The pattern point \( z_{KL} \) can be defined as the collection of all relevant features and labels: \(
z_{KL} = \left\{ \left(x[k]^{(n)}, y[l]^{(n)}\right) \mid k \in K, l \in L \right\}_{n=1}^N\)
\label{zkl}

\paragraph{Notations} We will not differentiate between random variables and their specific realizations if there is no risk of confusion. For any set of random variables \(X\), the joint probabilistic distribution functions (PDFs) under different models are denoted as \(\mathbb{P}(X)\), \(\mathbb{Q}(X)\), etc. Specifically, \(\mathbb{P}(\cdot)\) refers to the induced PDF in our proposed model, while \(\mathbb{Q}(X)\) and \(\mathbb{Q}_i(X)\), \(i \in \mathbb{N}\), represent variational distributions.
In the case of discrete random variables, we use generalized PDFs that may include Dirac delta functions. Unless specified otherwise, \(\mathbb{E}[X]\) refers to the expectation over all random variables in \(X\) with respect to \(\mathbb{P}(X)\). Expectations concerning other distributions, denoted as \(\mathbb{Q}(x)\), are written as \(\mathbb{E}_{\mathbb{Q}(X)}[X]\).

\paragraph{Information Bottleneck} In supervised learning-based unlearning tasks, the Information Bottleneck method provides a framework to efficiently forget specific instances while retaining the essential predictive capabilities of the model.
We denote $I(X,Y)$ as the mutual information between the random variables $X$ and $Y$ , which takes the form:
\begin{equation}
    I\left(X, Y\right)=\hat{P}\left(X, Y\right) \log \frac{\hat{P}\left(X, Y\right)}{\hat{P}\left(X\right) \hat{P}\left(Y\right)}=\int_X \int_Y p(x, y) \log \frac{p(x, y)}{p(x) p(y)} \mathrm{d} x \mathrm{~d} y.
\end{equation}
Details of Information Bottleneck are included in Additional Background in the Appendix \ref{IB}

\section{Unlearning Information Bottleneck}
In general, the unlearning information bottleneck (UIB) principle, derived from the traditional information bottleneck (IB) concept, mandates that the parameter-space representation \(\theta\) minimizes information from the dataset \(X_{\mathcal{D}\backslash\Delta \mathcal{D}}\) (Comprehensive Unlearning) while maximizing information relevant to the target \(Y_{\mathcal{D}\backslash\Delta \mathcal{D}}\) (Generalization, Efficiency and Performance). However, optimizing the most general UIB is challenging due to the potential correlations between data points. Typically, the i.i.d. assumption of data points facilitates the derivation of variational bounds and enables accurate estimation of these bounds to learn IB-based models. Nonetheless, in unlearning scenarios, such i.i.d. assumptions may not hold as data points could exhibit dependencies influenced by their initial inclusion in the training set or by shared features. To adequately handle such dependencies, we cannot simply isolate and remove individual data points without considering their impact on the entire dataset. In practice, we often deal with only a single realization of \(\mathbb{P}(\mathcal{D}\backslash\Delta \mathcal{D})\), indicating that only one instance of the dataset is available for training and subsequent unlearning. Therefore, approximating the optimal parameter-space representation \(\theta\) in the general UIB formulation becomes impractical without introducing additional assumptions to manage these dependencies effectively.
\paragraph{UIB Principle}
In unlearning scenarios, we rely on a widely accepted assumption for local dependence: given data from neighboring data points within a certain context of a data point \(z\) in the training dataset, the remaining data will be independent of \(z\). We apply this assumption to constrain the parameter space \(\Omega\) of optimal representations, leading to a more tractable UIB principle. Specifically, we assume that the optimal parameter-space representation \(\theta\) follows the dependence structure illustrated in the UIB formulation.

In this approach, \(\mathbb{P}(\theta | \mathcal{D}\backslash\Delta \mathcal{D})\) iterates through parameter-space representations to model correlations. In each iteration \(l\), the local dependence assumption is used: the representation will be refined by incorporating relevant neighboring information with respect to a structure \(\theta^{(l)}\). Here, \(\{\theta^{(l)}\}_{1 \leq l \leq L}\) is obtained by locally adjusting the original feature set, which ultimately controls the flow of information from the unlearned dataset. Finally, predictions will be made based on \(\theta^{(L)}\). Using this formulation, the objective reduces to the following optimization:
\begin{equation}
    \min_{\mathbb{P}(\theta^{(L)} | \mathcal{D}\backslash\Delta \mathcal{D}) \in \Omega} \mathrm{UIB}_{\beta}(\mathcal{D}\backslash\Delta \mathcal{D}; \theta^{(L)}) \triangleq \overset{\mathrm{def}}{=}[-I(Y_{\mathcal{D}\backslash\Delta \mathcal{D}}; \theta^{(L)}) + \beta I(X_{\mathcal{D}\backslash\Delta \mathcal{D}}; \theta^{(L)})],
\end{equation}
where \(\Omega\) characterizes the parameter-space conditional distribution of \(\theta^{(L)}\) given the data \(\mathcal{D}\backslash\Delta \mathcal{D}\), following the probabilistic dependence within the unlearning framework. Here, we need to optimize two sequences of distributions \(\mathbb{P}(\theta^{(l)} | \theta^{(l-1)}, \theta_\mathcal{R}^{(l)})\) and \(\mathbb{P}(\theta_\mathcal{R}^{(l)} | \theta^{(l-1)}, \mathcal{R})\) for \(l \in [L]\), which have local dependencies among data points and are therefore easier to parameterize and optimize. \(\mathcal{R}\) represents statistical independence relationships among the data points in each iteration, providing the framework necessary to guide parameter updates.

\subsection{Variational Bounds}
Even when following a reduced unlearning information bottleneck (UIB) principle and using appropriate parameterization for \(\mathbb{P}(\theta^{(l)} | \theta^{(l-1)}, \mathcal{R}^{(l)})\) and \(\mathbb{P}(\mathcal{R}^{(l)} | \theta^{(l-1)})\) for \(l \in [L]\), exact computation of \(I(Y_{\mathcal{D}\backslash\Delta \mathcal{D}}; \theta^{(L)})\) and \(I(X_{\mathcal{D}\backslash\Delta \mathcal{D}}; \theta^{(L)})\) remains intractable. Therefore, it is necessary to introduce variational bounds on these two terms to obtain an objective that can be optimized. Variational methods are often used to optimize models based on the traditional IB principle. However, we must derive these bounds carefully, given that the data points are correlated.

\textbf{Proposition 1} (Upper Bound of \(I(X_{\mathcal{D}\backslash\Delta \mathcal{D}}; \theta^{(L)})\)). We select two groups of indices, \(S_\theta\) and \(S_{\mathcal{R}}\), such that \(\mathcal{D} \perp \theta^{(L)} | \{\theta^{(l)}\}_{l \in S_\theta} \cup \{\mathcal{R}^{(l)}\}_{l \in S_{\mathcal{R}}}\) based on the probabilistic dependence. For any distributions \(\mathbb{Q}(\theta^{(l)})\) for \(l \in S_\theta\), and \(\mathbb{Q}(\mathcal{R}^{(l)})\) for \(l \in S_{\mathcal{R}}\), the bound is:
\begin{equation}
\begin{aligned}
&I(X_{\mathcal{D}\backslash\Delta \mathcal{D}}; \theta^{(L)}) \leq I(X_{\mathcal{D}\backslash\Delta \mathcal{D}}; \{\theta^{(l)}\}_{l \in S_\theta} \cup \{\mathcal{R}^{(l)}\}_{l \in S_{\mathcal{R}}}) \leq \sum_{l \in S_\theta} \text{UIB}_\theta^{(l)} + \sum_{l \in S_{\mathcal{R}}} \text{UIB}_\mathcal{R}^{(l)}, \text{ where } \\
&\text{UIB}_\theta^{(l)} = \mathbb{E}\left[\log \frac{\mathbb{P}(\theta^{(l)} | \theta^{(l-1)}, \mathcal{R}^{(l)})}{\mathbb{Q}(\theta^{(l)})}\right], \text{UIB}_\mathcal{R}^{(l)} = \mathbb{E}\left[\log \frac{\mathbb{P}(\mathcal{R}^{(l)} | \theta^{(l-1)})}{\mathbb{Q}(\mathcal{R}^{(l)})}\right].
\end{aligned}
\end{equation}
\paragraph{Proof} The first inequality follows directly from the data processing inequality and the Markov property \(\mathcal{D} \perp \theta^{(L)} | \{\theta^{(l)}\}_{l \in S_\theta} \cup \{\mathcal{R}^{(l)}\}_{l \in S_{\mathcal{R}}}\).

To prove the second inequality, we establish an ordering \(\prec\) on random variables in \(\{\theta^{(l)}\}_{l \in S_\theta} \cup \{\mathcal{R}^{(l)}\}_{l \in S_{\mathcal{R}}}\) such that:
1. For any two indices \(l, l'\), \(\theta^{(l)}, \mathcal{R}^{(l)} \prec \theta^{(l')}, \mathcal{R}^{(l')}\).
2. For a given index \(l\), \(\mathcal{R}^{(l)} \prec \theta^{(l)}\). 
Define the sets:
\begin{equation}
\begin{aligned}
G_\mathcal{R}^{(l)} & = \{\theta^{(l_1)}, \mathcal{R}^{(l_2)} \mid l_1 < l, l_2 < l, l_1 \in S_\theta, l_2 \in S_{\mathcal{R}}\}, \\
G_\theta^{(l)} & = \{\theta^{(l_1)}, \mathcal{R}^{(l_2)} \mid l_1 < l, l_2 \leq l, l_1 \in S_\theta, l_2 \in S_{\mathcal{R}}\}.
\end{aligned}
\end{equation}
Using this ordering, we decompose:
\[
I(X_{\mathcal{D}\backslash\Delta \mathcal{D}}; \{\theta^{(l)}\}_{l \in S_\theta} \cup \{\mathcal{R}^{(l)}\}_{l \in S_{\mathcal{R}}}) = \sum_{l \in S_{\mathcal{R}}} I(X_{\mathcal{D}\backslash\Delta \mathcal{D}}; \mathcal{R}^{(l)} | G_\mathcal{R}^{(l)}) + \sum_{l \in S_\theta} I(X_{\mathcal{D}\backslash\Delta \mathcal{D}}; \theta^{(l)} | G_\theta^{(l)}).
\]
Next, we bound each term:

1. For \(I(X_{\mathcal{D}\backslash\Delta \mathcal{D}}; \mathcal{R}^{(l)} | G_\mathcal{R}^{(l)})\):
\begin{equation}
I(X_{\mathcal{D}\backslash\Delta \mathcal{D}}; \mathcal{R}^{(l)} | G_\mathcal{R}^{(l)}) \leq I(X_{\mathcal{D}\backslash\Delta \mathcal{D}}, \theta^{(l-1)}; \mathcal{R}^{(l)} | G_\mathcal{R}^{(l)}) = \text{UIB}_\mathcal{R}^{(l)};
\label{eq:b1}
\end{equation}
2. For \(I(X_{\mathcal{D}\backslash\Delta \mathcal{D}}; \theta^{(l)} | G_\theta^{(l)})\):
\begin{equation}
I(X_{\mathcal{D}\backslash\Delta \mathcal{D}}; \theta^{(l)} | G_\theta^{(l)}) \leq I(X_{\mathcal{D}\backslash\Delta \mathcal{D}}, \theta^{(l-1)}, \mathcal{R}^{(l)}; \theta^{(l)} | G_\theta^{(l)}) = \text{UIB}_\theta^{(l)}.
\label{eq:b2}
\end{equation}
For equations \ref{eq:b1} and \ref{eq:b2}, we will give formal proof in Appendix \ref{pa}. In the decomposition and subsequent bounds above, (1) The inequalities leverage the basic properties of mutual information, allowing us to decompose and analyze the mutual information terms individually, and (2) The ordering \(\prec\) is crucial to ensure that each term considers dependencies in the correct hierarchical manner. For instance, conditioning on the sets \(G_\mathcal{R}^{(l)}\) and \(G_\theta^{(l)}\) effectively isolates relevant dependencies; (3) The terms \(\text{UIB}_\mathcal{R}^{(l)}\) and \(\text{UIB}_\theta^{(l)}\) are defined to capture the information bottleneck structure specific to each random variable set, ensuring that the resulting bounds provide meaningful estimates of information.

Therefore, the upper bound on \(I(X_{\mathcal{D}\backslash\Delta \mathcal{D}}; \theta^{(L)})\) can be expressed as a sum of the UIB terms for each hierarchical set \(S_\theta\) and \(S_{\mathcal{R}}\).
\[
I(X_{\mathcal{D}\backslash\Delta \mathcal{D}}; \theta^{(L)}) \leq \sum_{l \in S_\theta} \text{UIB}_\theta^{(l)} + \sum_{l \in S_{\mathcal{R}}} \text{UIB}_\mathcal{R}^{(l)}.
\]
Proposition 1 implies that we need to choose groups of random variables indexed by \(S_\theta\) and \(S_{\mathcal{R}}\) to ensure conditional independence between \(\mathcal{D}\) and \(\theta^{(L)}\). The indices sets \(S_\theta\) and \(S_{\mathcal{R}}\) that satisfy this condition have the following properties: (1) \(S_\theta \neq \emptyset\), and (2) if the highest index in \(S_\theta\) is \(l\), then \(S_{\mathcal{R}}\) should include all integers in \([l + 1, L]\).
\paragraph{Lemma 1} \citep{poole2019variational} For any two random variables $X_1, X_2$ and any function $g: g\left(X_1, X_2\right) \in \mathbb{R}$, we have
$$
I\left(X_1, X_2\right) \geq \mathbb{E}\left[g\left(X_1, X_2\right)\right]-\mathbb{E}_{\mathbb{P}\left(X_1\right) \mathbb{P}\left(X_2\right)}\left[\exp \left(g\left(X_1, X_2\right)-1\right)\right]
$$

We apply the above lemma for $\left(_{\mathcal{D}\backslash\Delta \mathcal{D}}, \theta^{(L)}\right)$ and plug in 
\[
g\left(Y_{\mathcal{D}\backslash\Delta \mathcal{D}}, \theta^{(L)}\right)=1+\log \frac{\prod_{v \in V} \mathbb{Q}_1\left(Y_{\mathcal{D}\backslash\Delta \mathcal{D},v} \mid \theta_{v}^{(L)}\right)}{\mathbb{Q}_2(Y_{\mathcal{D}\backslash\Delta \mathcal{D}})}. 
\]
For any conditional distribution \(\mathbb{Q}_1(Y_{\mathcal{D}\backslash\Delta \mathcal{D}} | \theta^{(L)})\) and marginal distribution \(\mathbb{Q}_2(Y_{\mathcal{D}\backslash\Delta \mathcal{D}})\), the \(I(Y_{\mathcal{D}\backslash\Delta \mathcal{D}}; \theta^{(L)})\) lower bound is:
\[
I(Y_{\mathcal{D}\backslash\Delta \mathcal{D}}; \theta^{(L)}) \geq \mathbb{E}\left[\log \frac{\mathbb{Q}_1(Y_{\mathcal{D}\backslash\Delta \mathcal{D}} | \theta^{(L)})}{\mathbb{Q}_2(Y_{\mathcal{D}\backslash\Delta \mathcal{D}})}\right] + C,
\]
where \(C\) is a constant dependent on prior distributions.

To apply UIB, we model \(\mathbb{P}(\mathcal{R}^{(l)} | \theta^{(l-1)})\) and \(\mathbb{P}(\theta^{(l)} | \theta^{(l-1)}, \mathcal{R}^{(l)})\). Then, we select appropriate variational distributions \(\mathbb{Q}(\theta^{(l)})\) and \(\mathbb{Q}(\mathcal{R}^{(l)})\) to estimate the corresponding \(\text{UIB}_{\theta}^{(l)}\) and \(\text{UIB}_{\mathcal{R}}^{(l)}\) for regularization. For lower-bound estimation, we define distributions \(\mathbb{Q}_1(Y | \theta^{(L)})\) and \(\mathbb{Q}_2(Y)\). Inserting these distributions into the UIB principle yields an upper bound on the objective function for optimization.
Note that any model parameterizing \(\mathbb{P}(\mathcal{R}^{(l)} | \theta^{(l-1)})\) and \(\mathbb{P}(\theta^{(l)} | \theta^{(l-1)}, \mathcal{R}^{(l)})\) can utilize UIB as an optimization objective. In the next subsection, we will introduce two specific instantiations of UIB that address unlearning in machine learning models.

\subsection{Implementing UIB}
The UIB framework can be effectively integrated into various machine learning models to enhance unlearning capabilities. For instance, in a scenario applying UIB to a machine unlearning task, we present the UIB-Influence Function. The base framework of these models is delineated through an illustrative algorithm, which includes different unlearning sampling methods further explained in subsequent algorithm descriptions. Each iteration within the UIB-Influence Function initially refines the parameter space using regularization techniques to derive $\mathcal{R}^{(l)}$, followed by updating the model parameters $\theta^{(l)}$ by considering influences from $\theta^{(l-1)}$ using the structured regularization term $\mathcal{R}^{(l)}$.

We design two distinct unlearning sampling algorithms, which utilize the categorical and Bernoulli distributions. In the categorical approach, regularization parameters act as the probabilities for the categorical distribution to sample and refine the parameter adjustments, extracting relevant information for efficient unlearning. We sample $k$ parameters with replacement from a predefined set for each parameter update, where the set includes parameters influenced within a threshold $\mathcal{T}$, which encapsulates the local-dependence assumption of the UIB principle enhancing the model's scalability.

To apply the Unlearning Information Bottleneck (UIB) principle across approximate unlearning methods, including influence functions, gradient ascent, and Fisher forgetting, we focus on how parameter estimates \(\theta^{(l)}\) change during the unlearning process. Here’s a step-by-step derivation that provides insights for incorporating influence functions and norms into UIB:

\paragraph{Implementing in Unlearning with an Influence Function}
where the inclusion of \(\mathcal{R}^{(l)}\) aligns the influence function within the UIB methodology.
The influence function approximates the impact of perturbing or removing a pattern point \(z_{KL} = \left\{ \left(x[k]^{(n)}, y[l]^{(n)}\right) \mid k \in K, l \in L \right\}_{n=1}^N\) (defined in Section. \ref{zkl}) from the dataset. Following the analogy and derivations presented in \citep{koh2017understanding}, we arrive at the following conclusion:
\[
\left.\mathcal{I}_{\text{up,params}}(z_{KL}) \stackrel{\text{def}}{=} \frac{d \hat{\theta}_{\epsilon, z_{KL}}}{d \epsilon}\right|_{\epsilon=0} = -H_{\hat{\theta}}^{-1} \nabla_{\theta} L(z_{KL}, \hat{\theta}),
\]
\(\mathcal{I}_{\text {up,params }}(z_{KL})\) represents influence function\citep{cook1980characterizations} approximately evaluates the influence of upweighting $z_{KL}$ on the parameters $\hat{\theta}$. Whereas $H_{\hat{\theta}}$ is the Hessian matrix of model parameters. 
In each unlearning iteration for parameter updates at level \( l \), we aim to minimize the mutual information between \(\theta^{(l)}\) and \( X_{\mathcal{D}\backslash\Delta \mathcal{D}} \). We express this objective as:
\[
\text{UIB}_{\theta}^{(l)} = \mathbb{E}\left[\log \frac{\mathbb{P}(\theta^{(l)} | \theta^{(l-1)}, \mathcal{R}^{(l)})}{\mathbb{Q}(\theta^{(l)})}\right].
\]
Incorporate the regularization term \(\mathcal{R}^{(l)}\) for parameter refinement:\(\mathcal{R}^{(l)} = \|z_{KL}, \theta^{(l)}\|^2\) 
, we add this constraint to the influence function approximation. Thus, the new objective becomes:
\[
\text{UIB}_{\theta}^{(l)} \approx \mathbb{E}_{z_{KL}}\left[\log \frac{\mathbb{P}(\hat{\theta}^{(l-1)} - \mathcal{I}_{\text{up,params}}(z_{KL}) | \theta^{(l-1)}, \|z_{KL}, \theta^{(l)}\|^2)}{\mathbb{Q}(\hat{\theta}^{(l-1)} - \mathcal{I}_{\text{up,params}}(z_{KL}))} \right].
\]

The influence function approximates the impact on parameter estimates when a pattern point is removed. Thus, the new estimate \(\hat{\theta}_{-z_{KL}}\) after removing a pattern point \( z_{KL} \) is \citep{warnecke2021machine}:
\[
\hat{\theta}_{-z_{KL}} \approx \theta - .\mathcal{I}_{\text {up,params }}(z_{KL}) = \theta-
.\frac{\partial \theta_{\epsilon, z \rightarrow z}^*}{\partial \epsilon}|_{e=0}\approx \theta-H_{\theta}^{-1}\left(\nabla_\theta \ell\left(\tilde{z}, \theta\right)-\nabla_\theta \ell\left(z, \theta\right)\right).
\]
To efficiently compute \(H_{\hat{\theta}}^{-1} v\) for influence function evaluation, explicit storage of \(H_{\hat{\theta}}\) could be avoided \citep{agarwal2017second}  and the iterative approximation is:
\[
\tilde{H}_0^{-1} v =v, \tilde{H}_j^{-1} v = v + \left(I - \nabla_{\theta}^2 L\left(z_i, \theta\right)\right) \tilde{H}_{j-1}^{-1} v,
\]
\begin{wrapfigure}{r}{0.58\textwidth}
\begin{minipage}{0.58\textwidth}
\begin{algorithm}[H]
\caption{Unlearning Information Bottleneck with IF}
\begin{algorithmic}
\small
\State \textbf{Input:} Unlearn Request $\Delta \mathcal{D}$, Model parameters $\theta$
\State \textbf{Output:} Updated model parameters $\hat{\theta}$

\For{each layer $l = 1$ to $L$} 
    \For{each pattern point $z_{KL}$ in $\Delta \mathcal{D}$}
        \State Compute influence function $\mathcal{I}_{\text{up,params}}(z_{KL})$
        \State Update parameters $\hat{\theta}^{(l)} \leftarrow \hat{\theta}^{(l-1)} - \mathcal{I}_{\text{up,params}}(z_{KL})$
        \State Regularize update: $\hat{\theta}^{(l)} \leftarrow \hat{\theta}^{(l)} - \lambda \cdot \text{reg}(\theta^{(l)}, z_{KL})$
        \State Approximate UIB term for $z_{KL}$ and update $\hat{\theta}^{(l)}$
    \EndFor
    \State Evaluate overall UIB for layer $l$
\EndFor
\end{algorithmic}
\end{algorithm}
\end{minipage}
\end{wrapfigure}
where $v$ is a vector of the same dimension as the model parameter $\theta$,  which serves as the input to the product operation of the Hessian matrix $H$. This technique enables efficient updates using Hessian-Vector-Products (HVPs) calculated directly from the gradient function, significantly reducing computational overhead in $\mathcal{O}(p)$ \citep{pearlmutter1994fast}.
% This line will wrap the text around the right side of the algorithm block.
\paragraph{Generalizing to Other Methods}
While influence functions offer one approximation method, others like Gradient Ascent \citep{graves2021amnesiac,thudi2022unrolling} and Fisher Forgetting \citep{golatkar2020eternal,becker2022evaluating} can also be incorporated using the UIB framework. This is achieved by ensuring the regularization term \(\mathcal{R}^{(l)}\) is well-defined, guiding the parameter updates while maintaining consistency with the UIB principle.
Thus, \(\text{UIB}_{\theta}^{(l)}\) provides a comprehensive way to approach approximate unlearning methods through careful integration of influence functions, norms, and other approximation techniques.
\section{Experiments}
Our experimental evaluation aims to determine the efficacy of our unlearning methods in effectively removing systematic patterns and biases while maintaining the robustness and accuracy of neural networks. Specifically, our experiments are designed to answer two key questions:
(1) Can UIB purge systematic patterns and biases effectively, ensuring unlearning efficacy?
(2) How does the adaptability of UIB manifest across different datasets and models, particularly regarding their ability to maintain high model performance post-unlearning?

\paragraph{Datasets and models.} Our experiments utilize four widely recognized datasets: MNIST\citep{deng2012mnist}, MNIST-C (Colored)\citep{addepalli2022feature}, CIFAR-10\citep{alex2009learning} and CIFAR-100\citep{alex2009learning}, adhering to the standard image classification framework (ResNet-18) \citep{he2016deep} and typical train-validation-test splits as established in prior studies \cite{liu2024model}. Unless specified otherwise, our experiments will focus on CIFAR-10 and ResNet-18. Dataset statistics and their specific splits are elaborated in Tab. \ref{tb:DS} of Appendix \ref{ap:DS}. 
\begin{wrapfigure}{l}{0.4\textwidth}
  \begin{minipage}{0.4\textwidth}
    \centering
    \includegraphics[width=\textwidth]{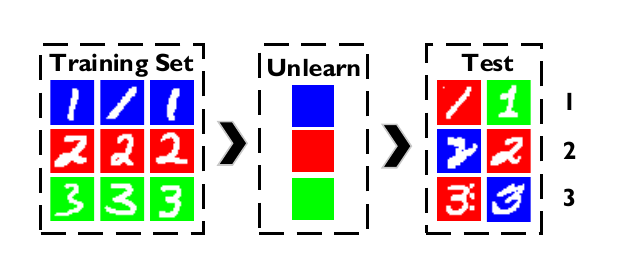}
    \captionsetup{font=small}
    \caption{Experiment Settings. The background colors are fixed for each number in the training set. The unlearning request is to forget all the biases (colors) the model learned from the training set. The correlation of bias and prediction and performance is tested on a test set of uniform and randomly matched colors and numbers.}
    \label{figE4}
  \end{minipage}
\end{wrapfigure}

\textbf{Unlearning settings and frameworks.} We focus on two unlearning scenarios: Systematic Patterns Unlearning and Random Data Points Unlearning. The total data we forget in both settings is limited to $10\%$ ($10\%$ of the total systematic patterns-related features and $10\%$ of the data points). We compare our UIB framework with Sparsity Settings (with 95\% sparsity rate) \citep{liu2024model} and the standard settings of respective approximate unlearning methods (Fine Tuning [FT]\citep{golatkar2020eternal,warnecke2021machine}, Scrubbing [SR]\citep{golatkar2020eternal, graves2021amnesiac} and Influence Function [IF]\citep{izzo2021approximate, warnecke2021machine, koh2017understanding}). We implement these methods following their official repositories. For Bias unlearning, we conduct experiments on an MNIST-C dataset, which we have partitioned into two distinct components: one containing RGB solid colors and the other featuring basic MNIST shapes. We artificially introduced significant biases, such as colors, as shown in Fig. \ref{figE4}, and tasked the unlearning model with removing these biases learned during training.

\textbf{Evaluation.} We evaluate the unlearning performance with Bias-Correlation (Correlation between biases and predictions), MIA-Efficacy (Membership Inference Attack \citep{song2019privacy, yeom2018privacy} Efficacy \citep{song2021systematic}, defined in Appendix \ref{MIA}, RIP (Relative Improvement Percentage, defined in Appendix \ref{RIP}), Testing F1 Score, and UT (Unlearning Time). Note that Bias-Correlation and MIA-Efficacy depict the unlearning \textit{efficacy}, Testing F1 Score characterizes \textit{generalizability} and \textit{accuracy}, and UT reflects \textit{efficiency}. We compute mean/std values of metrics and report std values as error bars in our analysis.
We maintain consistent architectural components for comparative models and vary parameters critical to the unlearning process, selecting the best-performing settings for each experimental scenario as detailed in Appendix \ref{ap:Hyper}.\label{exp}

\subsection{Experimental Results}
\label{res}
\begin{wrapfigure}{r}{0.49\textwidth}
  \begin{minipage}{0.49\textwidth}
    \centering
    \includegraphics[width=0.49\textwidth]{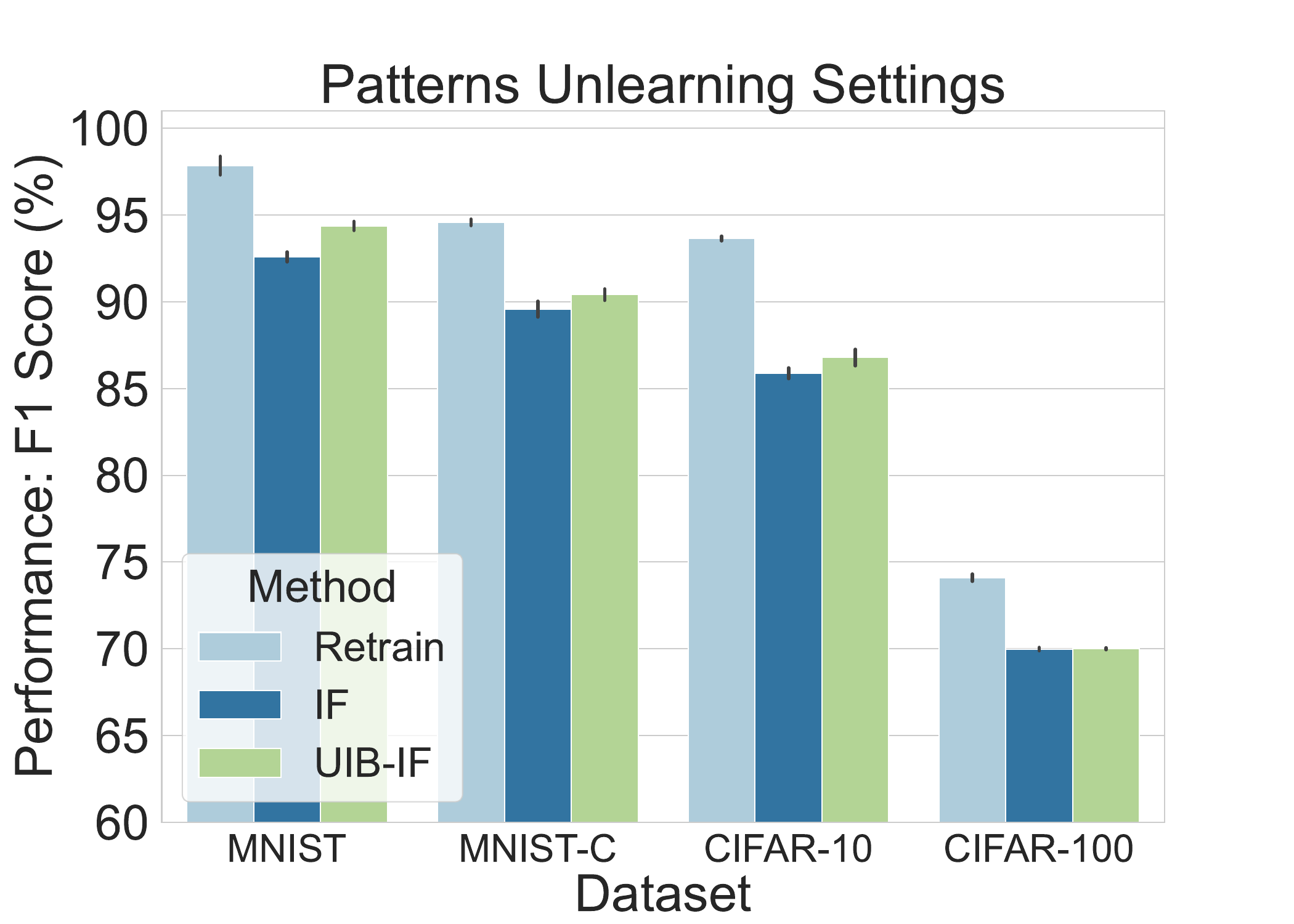}
    \includegraphics[width=0.49\textwidth]{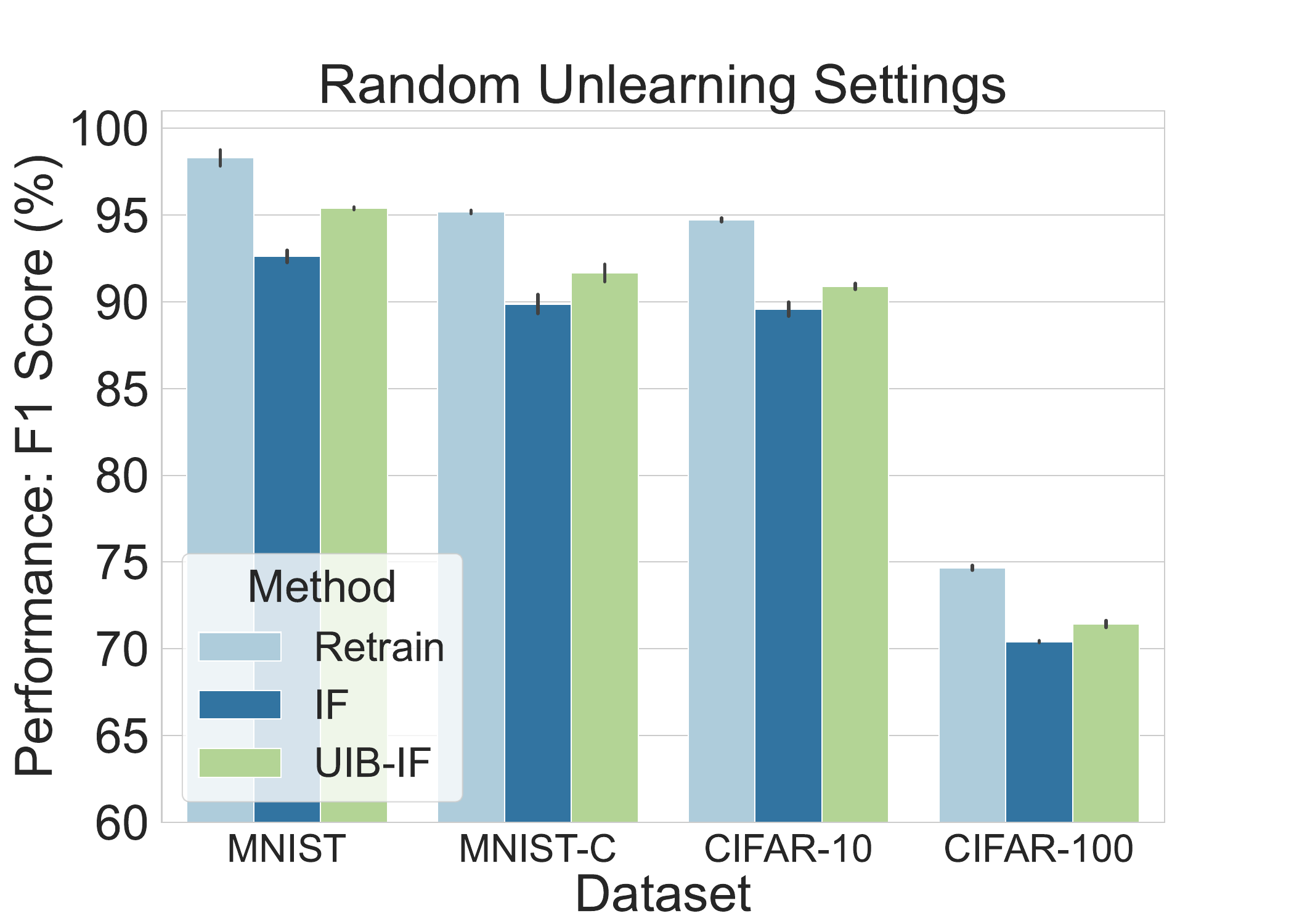}
    \captionsetup{font=small}
    \caption{Post-unlearning Accuracy of UIB. We compared Exact Unlearning, Approximate Unlearning and UIB's test F1 Score in System Patterns and Random Data Points Scenarios, respectively.}
    \label{figE1}
  \end{minipage}
\end{wrapfigure}

In Tab. \ref{ET1}, we conduct a thorough analysis of the Unlearning Information Bottleneck (UIB) method compared to standard unlearning and 95\%-sparsity methods across the ResNet-18 model trained on both MNIST and CIFAR-10 datasets. This comparative analysis highlights UIB's capacity to effectively mitigate biases while preserving model robustness and efficacy after unlearning processes.

\textbf{Post-Unlearning Performance.}
The UIB method consistently enhances unlearning efficacy and narrows the performance differential compared to the traditional retraining approach. This is particularly evident in the Testing F1 Scores, where UIB-IF secures $94.37\%$ on MNIST and $86.79\%$ on CIFAR-10 within the patterns and bias unlearning scenarios. It also achieves $95.39\%$ on MNIST and $90.89\%$ on CIFAR-10 for random data points unlearning settings, demonstrating UIB's remarkable adaptability and ability to maintain high model performance post-unlearning.   
Fig. \ref{figE1} illustrates the effect of integrating UIB on the performance of various datasets. 
Furthermore, when evaluating the sparsity setting, UIB shows a lesser degradation in Testing F1 Score than the standard method, indicating its effectiveness in handling systematic unlearning without significant loss in model accuracy. For instance, the Testing F1 Score under UIB-IF for CIFAR-10 is significantly better at $86.79\%$ compared to $82.07\%$ in the sparsity setting. This pattern is consistent across both datasets, emphasizing UIB's superior balance between unlearning efficacy and model performance retention.
\begin{wrapfigure}{r}{0.5\textwidth}
  \begin{minipage}{0.5\textwidth}
    \centering
    \includegraphics[width=0.49\textwidth]{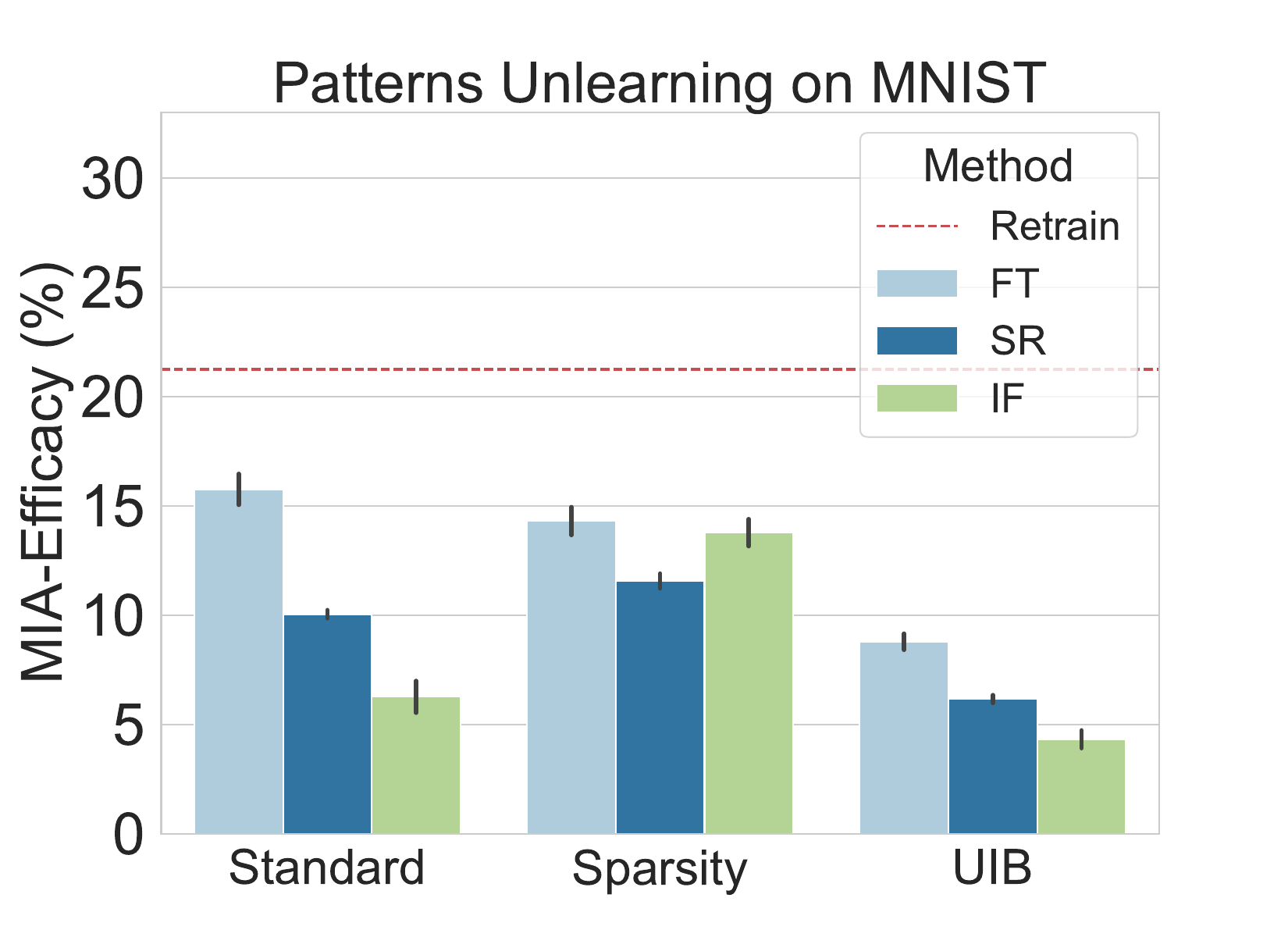}
    \includegraphics[width=0.49\textwidth]{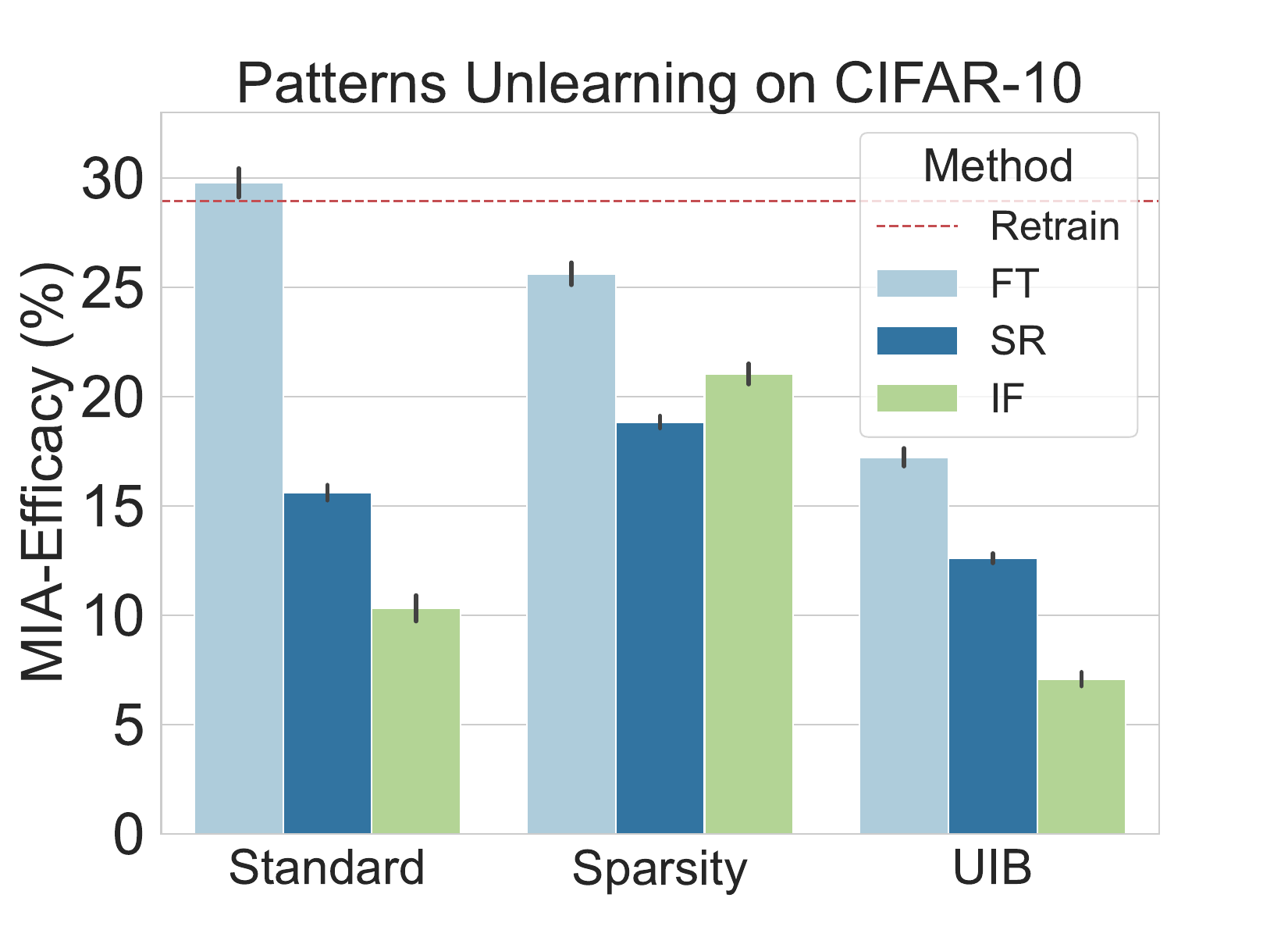}
    \captionsetup{font=small}
    \caption{MIA-Efficacy in Systematic Patterns Unlearning. We compared FT, SR, and IF methods in standard, sparsity, and UIB settings, respectively. The red line represents the baseline performance, lower is better.}
    \label{figEE}
  \end{minipage}
\end{wrapfigure}

Moreover, the impact of UIB on unlearning time (UT) is minimal, suggesting an efficient unlearning process. For example, the UT for UIB in the MNIST dataset is only slightly increased to $1.27$ minutes from $1.04$ minutes in the standard setting, a negligible trade-off for the gains in unlearning efficacy and model performance.
\begin{wrapfigure}{r}{0.5\textwidth}
  \begin{minipage}{0.5\textwidth}
    \centering
    \includegraphics[width=\textwidth]{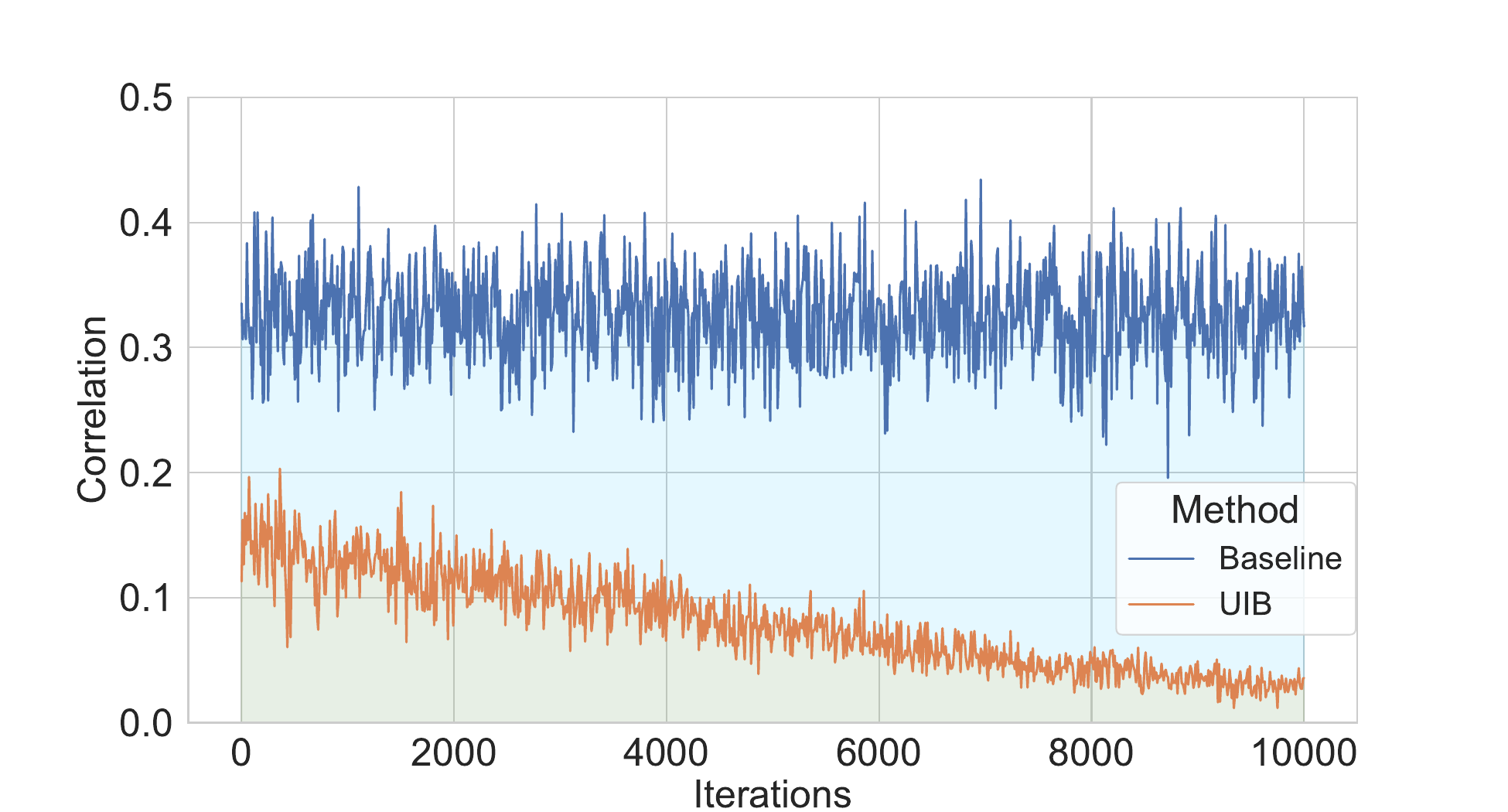}
    \captionsetup{font=small}
    \caption{Correlation of Unlearned Biases and Prediction. After each iteration of unlearning bias, the model was used to complete the test set's image classification task and calculate the correlation between image color and numeric labels.}
    \label{figE3}
  \end{minipage}
\end{wrapfigure}

\textbf{MIA-Efficacy and Privacy.} For the MNIST dataset under the UIB setting, there is a notable improvement in MIA-Efficacy down to $4.33\%$ in patterns and bias setting, demonstrating a more secured bias mitigation. Similarly, the CIFAR-10 results under UIB reveal an MIA-Efficacy of $7.08\%$, substantially better than the standard methods. These improvements suggest that UIB maintains the unlearning performance and enhances a security and privacy guarantee to the standard settings. In addition, UIB has better privacy-preserving advantages in both bias and random scenarios than sparsity setting, for instance, up to $13.95\%$ on IF with CIFAR-10 in bias scenario (shown as Fig. \ref{figEE}). For details and background about MIA-Efficacy, refer to Appendix \ref{MIA}

\textbf{Efficacy of Removing Patterns and Biases.}
In Fig. \ref{figE3}, we present the results of our experiment, where we compare the efficacy of different models in unlearning patterns consisting of systematic features across various iterations. The figure displays the correlation between unlearned patterns and bias and predictive target values in lighter shades. The results illustrate a notable reduction in correlation between the unlearning target and target over increasing iterations, contrary to what is observed in the baseline method. This indicates that UIB significantly mitigates the influence of systematic patterns and biases in the unlearning process.

\textbf{Additional results.} 
We have expanded our experimental scope to encompass the UIB framework across various datasets and architectures. In the UIB regularization strategy, we performed additional experiments in neural networks with other structures.
\begin{table}
\captionsetup{font=small}
\caption{Comparative analysis of UIB, sparsity, and standard unlearning methods on the ResNet-18 model trained on MNIST and CIFAR-10 datasets. Each metric is a mean $\pm$ standard deviation computed over 10 independent trials. The table aims to showcase the effectiveness of the UIB method in mitigating biases systematically compared to the baselines.}
\begin{adjustbox}{max width=\textwidth}
\begin{tabular}{ccc|cccc|cccc}
\toprule
\multicolumn{3}{c}{\textbf{Unlearning}} & \multicolumn{8}{c}{\textbf{Dataset}} \\
\midrule
\multirow{2}{*}{\textbf{Scenarios}} & \multirow{2}{*}{\textbf{Settings}} & \multirow{2}{*}{\textbf{Methods}} & \multicolumn{4}{c}{\textbf{MNIST}} & \multicolumn{4}{|c}{\textbf{CIFAR-10}} \\
\cmidrule{4-11}
& & & \textbf{RIP(\%)} & \textbf{MIA-Efficacy(\%)} & \textbf{Testing F1 Score(\%)} & \textbf{UT (min)}& \textbf{RIP(\%)} & \textbf{MIA-Efficacy(\%)} & \textbf{Testing F1 Score(\%)} & \textbf{UT (min)} \\
\hline
\multirow{10}{*}{Patterns and Biases} & \multirow{4}{*}{Standard} & Retrain & $0.00$ & $21.25 _{\pm 0.18}$ & $97.85 _{\pm 0.54}$ & 18.42& $0.00$ & $28.96 _{\pm 0.19}$ & $93.65 _{\pm 00.14}$ & 56.76\\ 
\cdashline{3-11}
& & FT & $-4.09$ & $15.76 _{\pm 0.71}$ & $93.85 _{\pm 0.37}$ & 1.31& $-9.93$ & $29.78 _{\pm 0.66}$ & $84.35 _{\pm 0.23}$ & 3.31\\ 
& & SR & $-7.51$ & $9.95 _{\pm 0.29}$ & $90.50 _{\pm 0.18}$ & 11.89& $-11.15$ & $15.61 _{\pm 0.35}$ & $83.21 _{\pm 0.11}$ & 37.61\\ 
& & IF & $-5.38$ & $6.27 _{\pm 0.73}$ & $92.59 _{\pm 0.29}$ & \textbf{1.04}& $-8.30$ & $10.32 _{\pm 0.59}$ & $85.88 _{\pm 0.31}$ & \textbf{3.16}\\ 
\cmidrule{2-11}
&\multirow{3}{*}{Sparsity} 
& FT & $-3.91$ & $14.31 _{\pm 0.64}$ & $94.02 _{\pm 0.51}$ & 1.87& $-9.71$ & $25.62 _{\pm 0.51}$ & $84.56 _{\pm 0.33}$ & 4.53\\ 
& & SR & $-10.21$ & $11.57 _{\pm 0.34}$ & $87.86 _{\pm 0.34}$ & 17.24& $-11.61$ & $18.84 _{\pm 0.28}$ & $82.78 _{\pm 0.20}$ & 51.27\\ 
& & IF & $-9.42$ & $13.78 _{\pm 0.62}$ & $88.63 _{\pm 0.48}$ & \textbf{1.76}& $-12.35$ & $21.03 _{\pm 0.47}$ & $82.07 _{\pm 0.53}$ & \textbf{4.28}\\ 
\cmidrule{2-11}
&\multirow{3}{*}{\textbf{UIB}} 
& \textbf{FT} & $-4.17$ & $8.79 _{\pm 0.37}$ & $93.77 _{\pm 0.42}$ & 1.59& $-8.95$ & $17.23 _{\pm 0.41}$ & $85.26 _{\pm 0.22}$ & 3.81\\ 
& & \textbf{SR} & $-7.71$ & $6.17 _{\pm 0.18}$ & $90.31 _{\pm 0.24}$ & 15.98& $-10.27$ & $12.62 _{\pm 0.22}$ & $84.03 _{\pm 0.17}$ & 48.56\\ 
& & \textbf{IF} & $-\textbf{3.56}$ & $\textbf{4.33} _{\pm \textbf{0.41}}$ & $\textbf{94.37} _{\pm \textbf{0.26}}$ & \textbf{1.27}& $-\textbf{7.33}$ & $\textbf{7.08} _{\pm \textbf{0.32}}$ & $\textbf{86.79} _{\pm \textbf{0.48}}$ & \textbf{3.49}\\ 
\midrule
\multirow{10}{*}{Random Data Points} & \multirow{4}{*}{Standard} & Retrain & $0.00$ & $13.67 _{\pm 0.12}$ & $98.29 _{\pm 0.46}$ & 16.98 & $0.12$ & $15.93 _{\pm 0.06}$ & $94.72 _{\pm 0.11}$ & 53.68\\ 
\cdashline{3-11}
& & FT & $-6.44$ & $14.97 _{\pm 0.69}$ & $91.96_{\pm 0.12}$ &1.27 & $-5.20$ & $16.12_{\pm 0.46}$ & $89.78_{\pm 0.09}$ & 3.19\\ 
& & SR & $-7.79$ & $10.37_{\pm 0.47}$ & $90.63 _{\pm 0.19}$ &10.78 & $-7.71$ & $11.27 _{\pm 0.31}$ & $87.42_{\pm 0.25}$ & 35.98\\ 
& & IF & $-5.77$ & $\textbf{5.04}_{\pm \textbf{0.56}}$ & $92.62 _{\pm 0.36}$ & \textbf{0.99}& $-5.43$ & $\textbf{5.73}_{\pm \textbf{0.29}}$ & $89.58_{\pm 0.41}$ & \textbf{3.04}\\ 
\cmidrule{2-11}
&\multirow{3}{*}{Sparsity} 
& FT & $-7.90$ & $14.28 _{\pm 0.33}$ & $90.53 _{\pm 0.24}$ &2.39 & $-5.60$ & $16.38_{\pm 0.09}$ & $89.42 _{\pm 0.14}$ & 4.36 \\ 
& & SR & $-11.46$ & $12.14 _{\pm 0.21}$ & $87.03 _{\pm 0.41}$ & 16.73 & $-8.16$ & $14.17_{\pm 0.22}$ & $86.99 _{\pm 0.32}$ & 46.23\\ 
& & IF & $-8.99$ & $13.41_{\pm 0.48}$ & $89.45_{\pm 0.39}$ & \textbf{1.65}& $-8.33$ & $14.76 _{\pm 0.39}$ & $86.83_{\pm 0.43}$ & \textbf{4.07}\\ 
\cmidrule{2-11}
&\multirow{3}{*}{\textbf{UIB}} 
& \textbf{FT} & $-6.00$ & $11.89_{\pm 0.51}$ & $92.39_{\pm 0.08}$ & 1.34& $-5.33$ & $13.97 _{\pm 0.29}$ & $89.67_{\pm 0.12}$ & 3.79 \\ 
& & \textbf{SR} & $-6.23$ & $9.76_{\pm 0.30}$ & $92.17_{\pm 0.16}$ &15.02 & $-7.47$ & $10.84 _{\pm 0.27}$ & $87.64_{\pm 0.19}$ & 45.88\\ 
& & \textbf{IF} & $-\textbf{2.95}$ & $ 10.14_{\pm 0.81}$ & $\textbf{95.39} _{\pm \textbf{0.08}}$ & \textbf{1.19}& $-\textbf{4.04}$ & $12.39 _{\pm 0.34}$ & $\textbf{90.89} _{\pm \textbf{0.18}}$ & \textbf{3.45}\\ 
\bottomrule
\end{tabular}
\end{adjustbox}
\label{ET1}
\end{table}
\section{Related Work}
\paragraph{Machine Unlearning.}
Machine unlearning techniques are differentiated into exact and approximate methods. Exact methods, like SISA \citep{bourtoule2021machine}, rigorously eliminate data influences by retraining models from scratch, catering to scenarios that demand robust data removal guarantees. Similarly, efforts by \citet{ginart2019making} on k-means clustering and \citet{karasuyama2010multiple} on support vector machines exemplify the exact unlearning approach. On the other hand, approximate methods, utilizing influence functions \citep{guo2019certified, izzo2021approximate}, aim to efficiently diminish the impact of data slated for unlearning while reducing computational overhead. These approaches are further refined by integrating differential privacy \citep{dwork2011differential}, which quantifies privacy post-unlearning to achieve an optimal balance between performance and privacy. Concurrently, the formulation of probabilistic unlearning models \citep{guo2019certified,ginart2019making,neel2021descent,ullah2021machine,sekhari2021remember}, particularly through the lens of differential privacy \citep{dwork2006our,dwork2011differential}, provides crucial error guarantees essential for defending against real-world adversaries and membership inference attacks. The exploration of data influence also informs practices in fair learning \citep{sattigeri2022fair,wang2022understanding}, transfer learning \citep{jain2023data}, and dataset pruning \citep{borsos2020coresets,yang2022dataset}, underscoring the extensive ramifications of understanding data's role in shaping model behavior and fostering more ethical and equitable machine learning methodologies.

\paragraph{Information Bottleneck.}
Information Bottleneck (IB) principle to unlearning tasks. The Deep Variational Information Bottleneck \citep{alemi2016deep} pioneered the application of the IB principle \citep{tishby2000information} to deep neural networks, enhancing the robustness of the learned representations. Other studies have extended the IB framework across diverse domains \citep{peng2018variational,higgins2017beta}, and structures \citep{wu2020graph,yu2020graph}, emphasizing the versatility and effectiveness of this approach. Distinct from these approaches, our work innovates on the application of the IB principle by focusing on the unlearning process of systematic patterns and bias, integrating the modeling of features, structures, and their interaction.

\section{Conclusion and Broader Impact}
This research introduces advancements in unlearning techniques, aiming to provide frameworks that support the removal or reduction of specific data influences in machine learning models. These methodologies will broadly benefit fields requiring stringent adherence to privacy regulations and data ethics, such as healthcare, finance, and social media.

\label{conclu}
\textbf{Limitations.}
The current framework has been evaluated under controlled conditions with a limited scope (independence assumptions and noiseless settings) of datasets and configurations. The performance and generalizability of the proposed methods should be generally robust across different domains or more complex data structures, except for particularly noisy scenes. 
If improperly implemented, unlearning techniques might lead to models that inaccurately reflect changes in data or fail to remove biases effectively. 

While our work on unlearning contributes to the safe and ethical use of AI, it is crucial to continue refining these techniques to handle the complex and often unpredictable nature of real-world data. Establishing rigorous standards for implementation and continuous monitoring of outcomes will be essential in mitigating the potential risks associated with these technologies.

\bibliographystyle{unsrtnat}
\bibliography{paper}
%%%%%%%%%%%%%%%%%%%%%%%%%%%%%%%%%%%%%%%%%%%%%%%%%%%%%%%%%%%%%%%%%%%%%%%%%%%%%%%
%%%%%%%%%%%%%%%%%%%%%%%%%%%%%%%%%%%%%%%%%%%%%%%%%%%%%%%%%%%%%%%%%%%%%%%%%%%%%%%
% APPENDIX
%%%%%%%%%%%%%%%%%%%%%%%%%%%%%%%%%%%%%%%%%%%%%%%%%%%%%%%%%%%%%%%%%%%%%%%%%%%%%%%
%%%%%%%%%%%%%%%%%%%%%%%%%%%%%%%%%%%%%%%%%%%%%%%%%%%%%%%%%%%%%%%%%%%%%%%%%%%%%%%
\newpage
\appendix
\onecolumn
\section{Additional Background}
\subsection{Information Bottleneck}
\label{IB}
In supervised learning-based unlearning tasks, the Information Bottleneck method provides a framework to efficiently forget specific instances while retaining the essential predictive capabilities of the model.
We denote $I(X,Y)$ as the mutual information between the random variables $X$ and $Y$ , which takes the form:
\begin{equation}
    I\left(X, Y\right)=\hat{P}\left(X, Y\right) \log \frac{\hat{P}\left(X, Y\right)}{\hat{P}\left(X\right) \hat{P}\left(Y\right)}=\int_X \int_Y p(x, y) \log \frac{p(x, y)}{p(x) p(y)} \mathrm{d} x \mathrm{~d} y.
\end{equation}
Given an unlearn request \(\Delta\mathcal{D}= (x_{\Delta D}, y_{\Delta D})\), the optimization objective leveraging the Information Bottleneck is defined as:
\begin{equation}
    \min_{\theta} \left[-I(Y_{D\backslash\Delta D}; \theta) + \beta I(X_{D\backslash\Delta D}; \theta)\right].
\end{equation}
This objective seeks to minimize the mutual information between the model parameters \(\theta\) and the outputs \(Y_{D\backslash\Delta D}\), while allowing some controlled dependency on the inputs \(X_{D\backslash\Delta D}\) scaled by a factor \(\beta\).

In parameter space, unlearning is an advanced approach that involves re-configuring the model's parameter space to efficiently forget specific data instances while retaining the overall structure and predictive capability of the model. The objective in parameter space unlearning is to minimize the mutual information between the unlearned dataset \(\Delta \mathcal{D}\) and the model's predictions on the remaining dataset \(\mathcal{D}\backslash\Delta \mathcal{D}\).
\subsection{Evaluation Metrics}
\paragraph{RIP}
\label{RIP}
To measure the degree to which an unlearning method reduces the F1 score compared to retraining and the magnitude of improvement, we calculate the Relative Improvement Percentage (RIP). This method provides an intuitive metric for assessing the effectiveness of unlearning methods in maintaining model performance:

The RIP can be defined as the degree of improvement in performance metrics of the unlearning method relative to the baseline methods. Specifically, RIP values can be calculated relative to complete retraining.
\begin{equation}
    RIP_{\text{retrain}} = \frac{F1_{\text{unlearn}} - F1_{\text{retrain}}}{F1_{\text{retrain}}} \times 100\%
\end{equation}

Where $F1_{\text{unlearn}}$ represents F1 score of the unlearning method and $F1_{\text{retrain}}$ represents F1 score of the completely retrained model. Negative $RIP$ indicates performance degradation of the unlearning method relative to the baseline. Zero $RIP$ indicates that the unlearning method is on par with the baseline in terms of performance, with no significant degradation.

\paragraph{MIA-Efficacy}
\label{MIA}
The Membership Inference Attack (MIA) Efficacy is another critical metric used to evaluate the success of our unlearning methods in removing specific data from a model. It is computed using a confidence-based prediction method to determine how well the model hides the presence of data points from the forgotten dataset, $\Delta D$. This metric reflects the privacy aspect of unlearning by measuring the ability of the model to prevent re-identification of the data points:

\[
MIA\text{-}Efficacy = 100\% - \left(\frac{\left|\{x \in \Delta D : \hat{y}_u(x) = 1\}\right|}{|\Delta D|} \times 100\%\right)
\]

where $\Delta D$ represents the dataset intended to be forgotten, $\hat{y}_u(x) = 1$ indicates that the instance $x$ is predicted by the unlearned model $f(D \setminus \Delta D)$ to still belong to the training set, and $|\Delta D|$ is the total number of instances in $\Delta D$.

A higher MIA-Efficacy value signifies that fewer data points from $\Delta D$ can be identified as belonging to the training set of the unlearned model, $f(D\backslash\Delta D)$. This indicates effective unlearning, as the model retains less information about the removed data, enhancing the privacy and security of the data subjects whose information was meant to be forgotten. This metric is especially important in scenarios where unlearning is used as a tool for complying with privacy regulations such as the GDPR, which mandates the right to erasure ("right to be forgotten").

\section{Expanded Proof for Bounds}
\label{pa}
Here we provide a detailed proof for the inequalities stated in Proposition 1 regarding the mutual information terms associated with the sets $S_\theta$ and $S_{\mathcal{R}}$ in the context of unlearning. 

\paragraph{Proof of Inequation \ref{eq:b1}}
Consider the inequality:
\[
I(X_{\mathcal{D}\backslash\Delta \mathcal{D}}; \mathcal{R}^{(l)} | G_\mathcal{R}^{(l)}) \leq I(X_{\mathcal{D}\backslash\Delta \mathcal{D}}, \theta^{(l-1)}; \mathcal{R}^{(l)} | G_\mathcal{R}^{(l)}).
\]
Using the chain rule for mutual information, we have:
\[
I(X_{\mathcal{D}\backslash\Delta \mathcal{D}}, \theta^{(l-1)}; \mathcal{R}^{(l)} | G_\mathcal{R}^{(l)}) = I(X_{\mathcal{D}\backslash\Delta \mathcal{D}}; \mathcal{R}^{(l)} | G_\mathcal{R}^{(l)}) + I(\theta^{(l-1)}; \mathcal{R}^{(l)} | X_{\mathcal{D}\backslash\Delta \mathcal{D}}, G_\mathcal{R}^{(l)}).
\]

In information theory, the definition of mutual information ensures that its value is always non-negative. Mutual information \(I(X; Y)\) measures the mutual dependence of the random variables \(X\) and \(Y\), or the amount of uncertainty in \(X\) that is reduced by knowing \(Y\). Mutual information is defined as:

\[
I(X; Y) = H(X) - H(X | Y)
\]

where \(H(X)\) is the entropy of \(X\), and \(H(X | Y)\) is the conditional entropy of \(X\) given \(Y\). Since entropy \(H(X)\) represents the uncertainty of the random variable, and \(H(X | Y)\) represents the uncertainty of \(X\) given \(Y\), thus \(H(X | Y) \leq H(X)\), this ensures that \(I(X; Y) \geq 0\).

In the proofs mentioned, the term \(I(\theta^{(l-1)}; \mathcal{R}^{(l)} | X_{\mathcal{D}\backslash\Delta \mathcal{D}}, G_\mathcal{R}^{(l)})\) represents the mutual information between \(\theta^{(l-1)}\) and \(\mathcal{R}^{(l)}\) given the knowledge of \(X_{\mathcal{D}\backslash\Delta \mathcal{D}}\) and the set \(G_\mathcal{R}^{(l)}\). This term is also non-negative, as it is defined based on the conditional entropy.

Thus, mathematically, the definition and properties of mutual information ensure that it is non-negative, which is a fundamental result in information theory.

The second term on the right-hand side is non-negative, ensuring the inequality holds.

\paragraph{Proof of Inequation \ref{eq:b2}}
Similarly, for the second term, we apply the chain rule:
\[
I(X_{\mathcal{D}\backslash\Delta \mathcal{D}}; \theta^{(l)} | G_\theta^{(l)}) \leq I(X_{\mathcal{D}\backslash\Delta \mathcal{D}}, \mathcal{R}^{(l)}; \theta^{(l)} | G_\theta^{(l)}),
\]
which expands to:
\[
I(X_{\mathcal{D}\backslash\Delta \mathcal{D}}, \mathcal{R}^{(l)}; \theta^{(l)} | G_\theta^{(l)}) = I(X_{\mathcal{D}\backslash\Delta \mathcal{D}}; \theta^{(l)} | G_\theta^{(l)}) + I(\mathcal{R}^{(l)}; \theta^{(l)} | X_{\mathcal{D}\backslash\Delta \mathcal{D}}, G_\theta^{(l)}).
\]
Here, the additivity of the mutual information and the non-negativity of the second term again justify the inequality.

These proofs utilize fundamental properties of mutual information, such as its non-negativity and the chain rule, to establish upper bounds in the structured unlearning setting, defined by the dependencies in $S_\theta$ and $S_{\mathcal{R}}$.

\section{Hyper-parameter and Experiment Details}
\subsection{Dataset Statistics (Tab. \ref{tb:DS})}
\label{ap:DS}
\begin{table}[ht]
\centering
\caption{Detailed information on the datasets used in the experiments.}
\begin{adjustbox}{max width=\textwidth}
\begin{tabular}{c|p{7cm}ccc}
\toprule
\centering
\textbf{Dataset} & \multicolumn{1}{c}{\textbf{Description}} & \textbf{Classes} & \textbf{Image Size} & \textbf{Total Images} \\
\midrule
\textbf{MNIST} & Handwritten digit images from 0 to 9. & 10 & 28x28 & 70,000 \\
\hline
\multirow{2}{*}{\textbf{MNIST-C}} & Colored variants of MNIST images, used for studying robustness against color-based perturbations. & \multirow{2}{*}{10} & \multirow{2}{*}{28x28} & \multirow{2}{*}{70,000} \\
\hline
\multirow{2}{*}{\textbf{CIFAR-10}} & Color images in 10 different classes, including animals and vehicles. & \multirow{2}{*}{10} & \multirow{2}{*}{32x32} & \multirow{2}{*}{60,000} \\
\hline
\multirow{2}{*}{\textbf{CIFAR-100}} & Color images in 100 classes grouped into 20 superclasses. & \multirow{2}{*}{100} & \multirow{2}{*}{32x32} & \multirow{2}{*}{60,000} \\
\bottomrule
\end{tabular}
\end{adjustbox}
\label{tb:DS}
\end{table}
\label{ap:Hyper}
\subsection{Error Bar Analysis}
To ensure the reliability and reproducibility of our experimental results, we have incorporated error bars representing the variability in our measurements. The error bars in all figures and tables represent the standard deviation of the mean (1-sigma), calculated over 10 independent trials. 

\textbf{Methodology for Calculating Error Bars}
The error bars were calculated using the bootstrap method, a robust statistical technique that resamples the original data to estimate the sampling distribution of a statistic. This method provides a good estimation of the standard deviation without the assumption of normal distribution of errors.

\textbf{Assumptions and Interpretation}
While our primary error reporting is based on the standard deviation, we also provide a 95\% Confidence Interval (CI) for key results. This interval was calculated assuming a t-distribution, which is a more appropriate assumption for the sample size of our experiments. This choice is due to the potentially non-normal distribution of errors, especially in cases involving deep learning models where the error distribution can be skewed or kurtotic.

\textbf{Factors Capturing Variability}
The variability captured by the error bars includes:
\begin{itemize}
    \item Variations across different train/test splits to account for changes in the data subsets used for training and evaluation.
    \item Initialization variability, considering the impact of different seeds on the stochastic nature of the training algorithms.
    \item Variations due to random drawing of hyper-parameters within predefined ranges, affecting model performance.
\end{itemize}

\textbf{Specific Metrics and Graphical Representations}
All error bars shown in plots and tables, such as those in Figure \ref{figE3} and Table \ref{ET1}, denote the standard deviation around the mean of measured metrics, such as F1 scores, MIA-Efficacy, and RIP values. Care has been taken to ensure that no error bars suggest impossible values, such as negative percentages. In cases of asymmetric error distributions, we adjusted the error bars to ensure all possible values are within logical bounds.

\textbf{Computational Resources}
The experiments were conducted using high-performance clusters equipped with A100 80G GPUs, 3 CPU cores, and 40G of memory. This setup was sufficient for the experiments conducted, including those with larger datasets like CIFAR-100. Computational requirements and hyper-parameter settings for each experiment are documented in Appendix B, ensuring transparency and reproducibility of our results.

\subsection{Hyper-parameter and Configurations (Tab.\ref{l})}
% We summarize the Hyper-parameter and model configurations in Tab. \ref{l}.
\begin{table}
\caption{Hyperparameter scope UIB-IF Implementation in Section \ref{exp}}
\centering
\begin{adjustbox}{{max width=\textwidth}}
\begin{tabular}{l|c|c}
\toprule
\multicolumn{1}{c}{\textbf{Hyperparameters}} & \multicolumn{1}{c}{\textbf{Value/Search space}} & \multicolumn{1}{c}{\textbf{Type}} \\
\midrule
$S_{\mathcal{R}}$ & {$[L]$} & Fixed \\
$S_\theta$ & $\{L-1\}$ & Fixed \\
Batch Size for MNIST & 256 & Fixed \\
Batch Size for MNIST-C & 256 & Fixed \\
Batch Size for CIFAR-10 & 256 & Fixed \\
Batch Size for CIFAR-100 & 256 & Fixed \\
$\beta$ & $\{0.1,0.01, 0.05\}$ & Choice \\
$\tau$ & $\{0.05,0.1,1\}$ & Choice \\
\bottomrule
\end{tabular}
\end{adjustbox}
\label{l}
\end{table}

\end{document}